\documentclass{article}
\usepackage{spconf,amsmath,graphicx}
\usepackage{amssymb,amsfonts}
\usepackage{algorithmic}
\usepackage{algorithm}
\usepackage{textcomp}
\usepackage{cleveref}
\usepackage{color}
\usepackage{multirow}

\usepackage{amsthm}
\usepackage{enumerate}
\usepackage[mathscr]{euscript}
\usepackage{url}
\usepackage[all]{xy}
\usepackage{booktabs}
\usepackage{graphicx}
\usepackage{epstopdf}
\usepackage[caption=false]{subfig}

\def\BibTeX{{\rm B\kern-.05em{\sc i\kern-.025em b}\kern-.08em
    T\kern-.1667em\lower.7ex\hbox{E}\kern-.125emX}}

\definecolor{darkred}{rgb}{0.7,0,0}
\definecolor{darkgreen}{rgb}{0,0.47,0}
\definecolor{purple}{rgb}{0.6,0,0.5}

\usepackage[mathscr]{euscript}
\usepackage{url}

\newcommand{\tens}[1]{\boldsymbol{\mathcal{#1}}}
\newcommand{\tenselem}[1]{\mathcal{#1}}

\newcommand{\matr}[1]{\boldsymbol{#1}}

\newcommand{\vect}[1]{\boldsymbol{#1}}
\newcommand{\set}[1]{\mathscr{#1}}
\newcommand{\T}{{\sf T}}        

\makeatletter
\newcommand{\rank}[1]{\mathop{\operator@font rank}\{#1\}}
\newcommand{\colrank}[1]{\mathop{\operator@font colrank}\{#1\}}
\newcommand{\krank}[1]{\mathop{\operator@font krank}\{#1\}}
\newcommand{\srank}[1]{\mathop{\operator@font srank}\{#1\}}
\newcommand{\trace}[1]{\mathop{\operator@font trace}\{#1\}}
\newcommand{\Diag}[1]{\mathop{\operator@font Diag}\{#1\}}    
\newcommand{\diag}[1]{\mathop{\operator@font diag}\{#1\}}    
\newcommand{\Span}[1]{\mathop{\operator@font Span}\{#1\}}    
\newcommand{\argmin}{\mathop{\operator@font argmin}}

\newcommand{\offdiag}[1]{\mathop{\operator@font offdiag}\{#1\}}    
\newcommand{\Proj}[2]{\mathop{\operator@font Proj_{#1}}{#2}}
\newcommand{\ProjGrad}[2]{\mathop{{\operator@font Proj} \nabla}#1(#2)}
\makeatother

\newcommand{\eqdef}{\stackrel{\sf def}{=}}
\newcommand{\RR}{\mathbb{R}}

\title{Point cloud denoising based on tensor Tucker decomposition}

\name{Jianze Li$^{\dagger}$ \qquad  Xiao-Ping Zhang$^{\ddagger}$ \qquad Tuan Tran$^{\ddagger}$\thanks{The work was supported in part by Shenzhen Peacock Plan under Grant No. KQTD2015033114415450 and the Natural
Sciences and Engineering Research Council of Canada
(NSERC, Grant No. RGPIN239031)}}
\address{$^{\dagger}$ Shenzhen Research Institute of Big Data, The Chinese University of Hong Kong, Shenzhen, China\\
$^{\ddagger}$ Electrical, Computer and Biomedical Engineering, Ryerson University, Toronto, ON, Canada\\
lijianze@gmail.com\qquad xzhang@ee.ryerson.ca\qquad tuan.maths@gmail.com}

\begin{document}
%
\maketitle
\begin{abstract}
In this paper, we propose a new algorithm for point cloud denoising based on the tensor Tucker decomposition. We first represent the local surface patches of a noisy point cloud to be matrices by their distances to a reference point, and stack the similar patch matrices to be a 3rd order tensor. Then we use the Tucker decomposition to compress this patch tensor to be a core tensor of smaller size. We consider this core tensor as the frequency domain and remove the noise by manipulating the hard thresholding. Finally, all the fibers of the denoised patch tensor are placed back, and the average is taken if there are more than one estimators overlapped. The experimental evaluation shows that the proposed algorithm outperforms the state-of-the-art graph Laplacian regularized (GLR) algorithm when the Gaussian noise is high ($\sigma=0.1$), and the GLR algorithm is better in lower noise cases ($\sigma=0.04, 0.05, 0.08$).
\end{abstract}
\section{Introduction}
\label{sec:intro}

In recent years, the low-cost and high-resolution scanners of point cloud are becoming available, and have been promoting the wide applications of point cloud processing in various areas, e.g., \emph{remote sensing} \cite{li2015hyperspectral}, \emph{cultural heritage} \cite{lozes2015pde} and \emph{geographic information system} \cite{sugimoto2017trends}. However, because of the physical constraints, the raw point cloud data is always corrupted with noises, which has made the denoising an important step for further processing in point cloud. In fact, the denoising problem is also a popular topic in signal processing field \cite{zhang1998thresholding,zhang1998thresholdingB,zhang1999thresholding,zhang2001thresholding,zhang2001thresholdingB,beheshti2010noise}.

Up to now, many different types of algorithms have been developed for point cloud denoising \cite{han2017review}, which can be classified as four categories \cite{zeng20183d}: Moving least squares (MLS)-based, Locally
optimal projection (LOP)-based, Sparsity-based and Non-local algorithms. 
Now we briefly introduce the \emph{algebraic point
set surfaces} (APSS) \cite{guennebaud2007algebraic} and \emph{robust
implicit MLS} (RIMLS) \cite{oztireli2009feature} algorithms in MLS-based category, and the \emph{graph Laplacian regularized} (GLR) \cite{zeng20183d} algorithm in Non-local category. 
In fact, the APSS and RIMLS algorithms are to approximate the smooth surface based on the local reference domains of the noisy points, and then determine the true positions by the resulting surface. The GLR algorithm is based on the assumption that the surface patches in a point cloud lie on a manifold of low dimension, which was earlier studied in the \emph{low-dimensional manifold
model} (LDMM) \cite{osher2017low} algorithm for image processing.

As more and more real world data can be represented as the tensor form, e.g., \emph{video} \cite{gui2017video}, \emph{hyperspectral image} \cite{li2015hyperspectral}, tensor decomposition has become a popular tool to solve many signal processing problems \cite{kolda2009tensor,comon2014tensors}, e.g., \emph{image denoising} \cite{rajwade2013image}, \emph{graph signal processing} (GSP) \cite{shahid2017scalable,gavili2017shift}. 
As one of the most important transformations in the tensor field, Tucker decomposition \cite{kolda2009tensor,tucker1966some} is to transform a tensor into a \emph{core tensor} of smaller size by a set of column orthogonal matrices.
In fact, it can be understood as a higher order version of the \emph{principal component analysis} (PCA).

In this paper,
we try to solve the point cloud denoising problem by a tensor approach.
We first construct a 3rd order tensor based on the similarity between different local surface patches in a point cloud.
Then we use the Tucker decomposition to compress the \emph{patch tensor}, and take the core tensor as the frequency domain.
In fact, this process has removed some noise, similar to the PCA case. For better denoising performance, we continue to use the hard thresholding on the core tensor to remove more noise. Finally, we place back the fibers\footnote{the fiber of a tensor is defined by fixing every index but one.}, and take the average if there are more than one estimators overlapped.
This algorithm belongs to Non-local category, as it is based on the similarity between local surface patches.
The main contribution of this paper is the formulation of the point cloud denoising problem from a tensor point of view, and combining the Tucker decomposition and hard thresholding to solve it. It is shown by the experiments that the proposed algorithm outperforms the state-of-the-art GLR algorithm when the Gaussian noise is high ($\sigma=$0.1). To the best of our knowledge, this is the first time applying Tucker decomposition to the point cloud denoising problem.

This paper is organized as follows. In \Cref{sec-problem}, we formulate the point cloud denoising problem, and propose a Tucker decomposition based algorithm to solve it. In \Cref{sec-experiment}, we conduct some experiments to evaluate the performance of the proposed algorithm, and discuss the results. \Cref{sec-conclusion} includes the conclusion and future work.


Now we give some notations before further discussion. 
In this paper, we denote $\RR^{m\times n\times p}\eqdef\RR^{m}\otimes\RR^{n}\otimes\RR^{p}$
to be the linear space of 3rd order real tensors.
We denote by $\|\cdot\|$ the Frobenius norm of a tensor or a matrix,
or the Euclidean norm of a vector.
Tensors, matrices, and vectors,  will be respectively denoted with bold calligraphic letters, e.g. $\tens{A}$, with bold uppercase letters, e.g. $\matr{X}$, and with bold lowercase letters, e.g. $\vect{v}$; corresponding entries will be denoted by $\tenselem{A}_{ijk}$, $X_{ij}$, and $v_i$.
Let $\tens{A}\in\RR^{m\times n\times p}$ be a 3rd order tensor, and $\matr{X}\in\RR^{q\times m}$ be a matrix.
We follow the definitions and notations in \cite{kolda2009tensor}, e.g., the \emph{1-mode product} is
\begin{equation*}
(\tens{A}\times_{1}\matr{X})_{ijk} \eqdef \sum_{l=1}^{m}\tenselem{A}_{ljk}X_{il}.
\end{equation*}

\section{Point cloud denoising based on Tucker decomposition}\label{sec-problem}

\subsection{Problem formulation}

Assume that $\set{V}=\{\vect{v}_{i}\}_{i=1}^{\rm N}$ is a noisy point cloud, i.e., a set of unstructured spatial points, 
and $\matr{V}=[\vect{v}_{1},\cdots,\vect{v}_{N}]^{\T}\in\RR^{N\times 3}$ is the corresponding position matrix satisfying
\begin{equation}\label{eq-denois-problem}
\matr{V} = \matr{W} + \matr{E},
\end{equation}
where $\matr{W}$ is the true position matrix and $\matr{E}$ is a Gaussian noise with zero mean and standard deviation $\sigma$. In this paper, we study the \emph{point cloud denoising} problem to find $\matr{W}$.

To solve this problem, we first represent the surface patches of a point cloud as matrices of the same size, and stack the similar \emph{patch matrices} to a 3rd order \emph{patch tensor} $\tens{A}\in\RR^{m\times n\times p}$, based on the similarity between different local surface patches.
Suppose that $r_1< m, r_2< n, r_3< p$. 
Then, we formulate problem \eqref{eq-denois-problem} to be the following Tucker decomposition \cite{kolda2009tensor,tucker1966some} problem
\begin{equation}\label{eq-tucker-problem}
\min \frac{1}{2}\|\tens{A}-\tens{C}\times_{1}\matr{P}\times_{1}\matr{Q}\times_{1}\matr{R}\|^2,
\end{equation}
where $\tens{C}\in\RR^{r_1\times r_2\times r_3}$ is the \emph{core tensor}, and $\matr{P}, \matr{Q},\matr{R}$ are the \emph{factor matrices}.
This is in fact the higher order PCA to compress $\tens{A}$ to be of smaller size. 
This problem can be solved efficiently by the \emph{higher order orthogonal iterations} (HOOI) \cite{de2000best} method.
In this paper, we use the HOOI algorithm to solve problem \eqref{eq-tucker-problem}, and understand the core tensor as the frequency domain similar to the PCA case.
Then by the hard thresholding on the core tensor, we remove the noise of the point cloud. 

\subsection{Tucker decomposition based algorithm}

In this subsection, we mainly develop the \emph{Tucker decomposition based point cloud denoising} (TUDE) algorithm. All the details of this algorithm are summarized in Algorithm 1.

\subsubsection{Determining the patch matrices}

We first choose a subset $\set{S}\subseteq\set{V}$ as the set of \emph{seed points} by the downsampling method, which can make the seed points sampled uniformly.
For each seed point $\vect{s}\in\set{S}$,
we choose the $K$ nearest points in $\set{V}$, and sort them to be a \emph{patch matrix} in $\RR^{K\times 3}$ based on their Euclidean distances to $\vect{s}$. The $\set{S}$ should be large enough to guarantee that the union of all the points in patch matrices can cover $\set{V}$.


\subsubsection{Determining the groups of similar patch matrices}

To make the similarity between patch matrices be
rotation invariant,
we use the distance $d(\matr{X},\matr{Y})$ based on the \emph{iterative closest point (ICP)} cost function, as in \cite{rosman2013patch}.

Given a thresholding value $\delta_{\rm sim}>0$, for each patch matrix $\matr{X}\in\RR^{K\times 3}$, we find all the patch matrices $\matr{Y}$ satisfying that
the average distance $d(\matr{X},\matr{Y})/3K$ is smaller than $\delta_{\rm sim}$. 
In the process of solving the ICP problem, we find a transformation on $\matr{Y}$. Then we put all of such transformed patch matrices in a group.
 In other words, the average distance between each patch matrice and the reference one in the same group will be always smaller than $\delta_{\rm sim}$.
 To guarantee the speed, we set a search region N$_{\rm reg}$ for this searching process.

\subsubsection{Patch tensor denoising}

For each group of similar patch matrices $\{\matr{X}_{p}\}_{p=1}^{M}\subseteq\RR^{K\times 3}$, we stack them together to be a 3rd order \emph{patch tensor} $\tens{A}\in\RR^{K\times 3\times M}$. Then we calculate the Tucker decomposition \eqref{eq-tucker-problem} of $\tens{A}$ by the HOOI algorithm, and manipulate the hard thresholding on the core tensor. We keep the entries with absolute value larger than the largest absolute value multiplied by $\delta_{\rm thre}$, and eliminate other entries.
Then, by the inverse transformation, we get the denoised tensor $\tens{A}_{\ast}\in\RR^{K\times 3\times M}$.

\subsubsection{Aggregation}

This step is to place all the fibers of denoised tensors $\tens{A}_{\ast}$ back to the original positions. On each patch matrix of the denoised tensors, we first make the inverse transformations of that determined in the process of solving the ICP problem. Then we place back all the patch matrices.
It is highly possible that we get more than one estimators for one original patch matrix. In this case, we take the average of these overlapped ones. Finally, we place back all the row vectors of patch matrices. It is also possible that there are several estimators for one point position, when a point appears simultaneously in many patch matrices, and we take the average similarly.

\begin{algorithm}[H]\label{alg-main}
	\begin{algorithmic}
		\renewcommand{\algorithmicrequire}{\textbf{Input:}}
		\renewcommand{\algorithmicensure}{\textbf{Output:}}
		\REQUIRE Noisy point cloud $\set{V}$;
			\STATE $K>0$, the number of points in each patch matrix;
			\STATE $\delta_{\rm sim}>0$, for finding similar patch matrices;
			\STATE N$_{\rm reg}>0$, the search region;
			\STATE $(r_1,r_2,r_3)$, the size of core tensor;
			\STATE $\delta_{\rm thre}>0$, for the hard thresholding.
		\ENSURE  Denoised point cloud $\set{V}_{\rm de}$.
		\\ \textit{Initialisation}
		\STATE Represent $\set{V}$ as a position matrix $\matr{V}$;
		\\ \textit{Phase I}: Determining the patch matrices.
		\STATE Choose a set of seed points $\set{S}\subseteq\set{V}$ by the downsampling method;
		\FOR {each seed point $\vect{s}\in\set{S}$}
		\STATE Choose the $K$ nearest points in $\set{V}$;
		\STATE Sort them to be a patch matrix by the distances to $\vect{s}$.
		\ENDFOR
		\\ \textit{Phase II}: Determining the groups of similar patch matrices.
		\FOR {each patch matrix $\matr{X}$}
		\STATE Find the group of similar patch matrices $\matr{Y}$.
		\ENDFOR
		\\ \textit{Phase III}: Patch tensor denoising.
		\FOR {each group of similar patch matrices}
		\STATE Stack them together to be a patch tensor $\tens{A}\in\RR^{K\times 3\times M}$;
		\IF {the size of $\tens{A}$ is greater than $(r_1,r_2,r_3)$}
		\STATE Calculate the Tucker decomposition \eqref{eq-tucker-problem} of $\tens{A}$;
		\STATE Manipulate the hard thresholding on the core tensor with parameter $\delta_{\rm thre}$;
		\STATE Inverse the transformations.
		\ENDIF
		\ENDFOR
		\\ \textit{Phase IV}: Aggregation.
		\STATE Place back all the fibers of denoised tensors;
		\STATE Take the average if there are more than one estimators.
	\end{algorithmic}
	\caption{Tucker decomposition based algorithm}
\end{algorithm}

\section{Experimental evaluation}\label{sec-experiment}

In this section, we compare the proposed TUDE algorithm with some existing algorithms: the APSS \cite{guennebaud2007algebraic}, RIMLS \cite{oztireli2009feature}, 
and GLR \cite{zeng20183d} algorithms.
The APSS and RIMLS algorithms are implemented with MeshLab software \cite{cignoni2008meshlab}. 
The GLR algorithm is implemented with Matlab.
The TUDE algorithm is implemented with Python.

\subsection{Experimental setup}

We use the  \emph{Gargoyle}, \emph{DC} and \emph{Daratech} point cloud models from \cite{rosman2013patch,mattei2017point,zeng20183d} to conduct experiments. The numbers of points and the numbers of seed points after the downsampling process for these three models are 58611, 56645, 32003 and 28361, 27496, 15475, respectively. 
These models are added the Gaussian noises with $\sigma={\rm 0.04, 0.05, 0.08, 0.1}$, respectively.
We use the \emph{mean square error} (MSE) in \cite{zeng20183d} to evaluate the denoising performance.


In the proposed TUDE algorithm, we always set $\delta_{\rm sim}=1$, N$_{\rm reg}=20$, $(r_1,r_2,r_3) = (3,3,3)$ and $\delta_{\rm thre}=0.1$. We set $K=19, 21, 26, 35$ for $\sigma={\rm 0.04, 0.05, 0.08, 0.1}$, respectively.
In the GLR algorithm, we set $r=12$ for $\sigma={\rm 0.04, 0.05}$, $r=13$ for $\sigma={\rm 0.08}$, and $r=14$ for $\sigma={\rm 0.1}$, as we find this setting can generally get the best results.
In APSS and RIMLS algorithms, we set the MSL filter scale to be $4, 5, 6$, and choose the best result.



\subsection{Experimental results}

\begin{figure}[tbhp]
	\centering
	\subfloat[]{\includegraphics[width=0.25\textwidth]{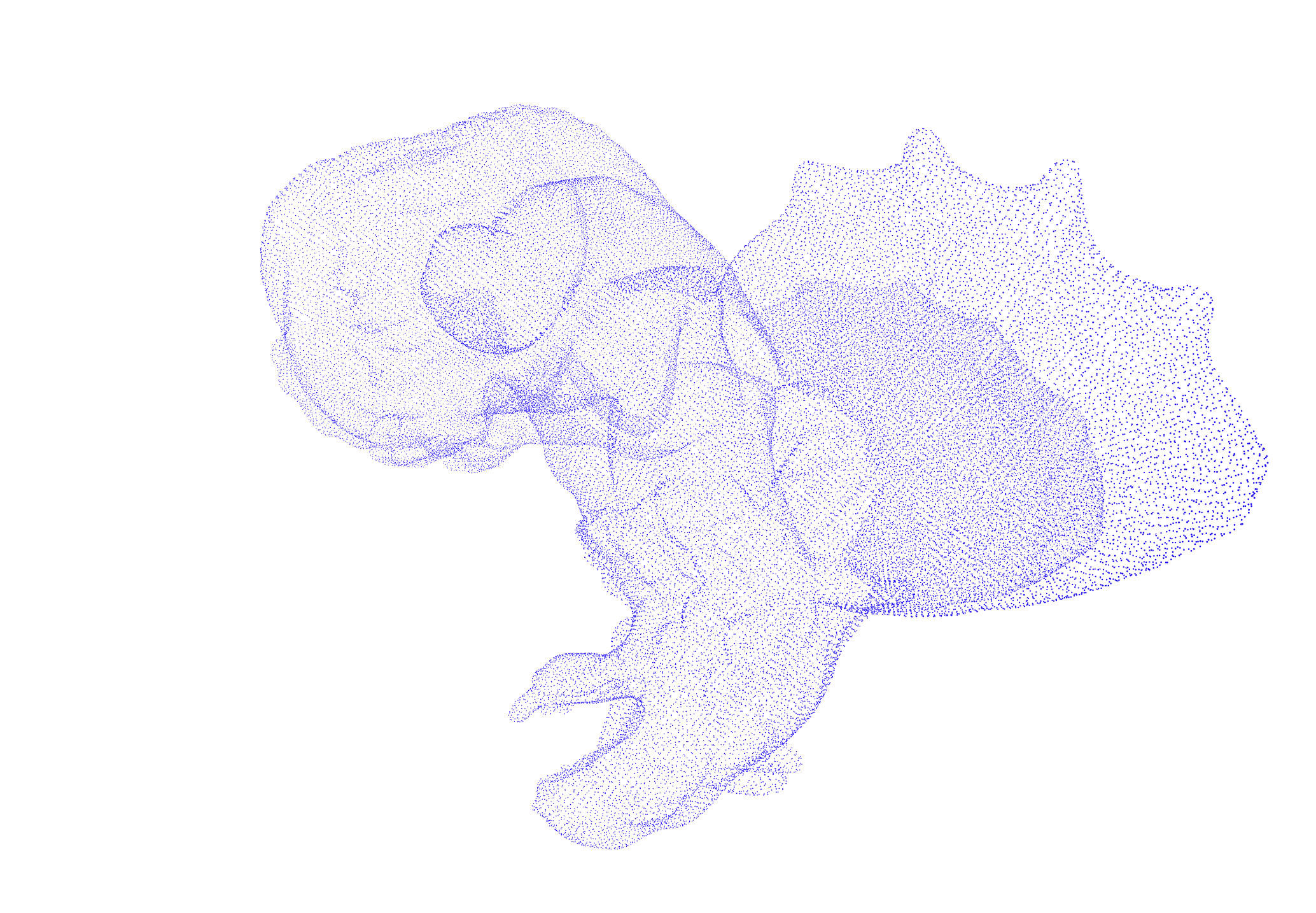}}
	\subfloat[]{\includegraphics[width=0.25\textwidth]{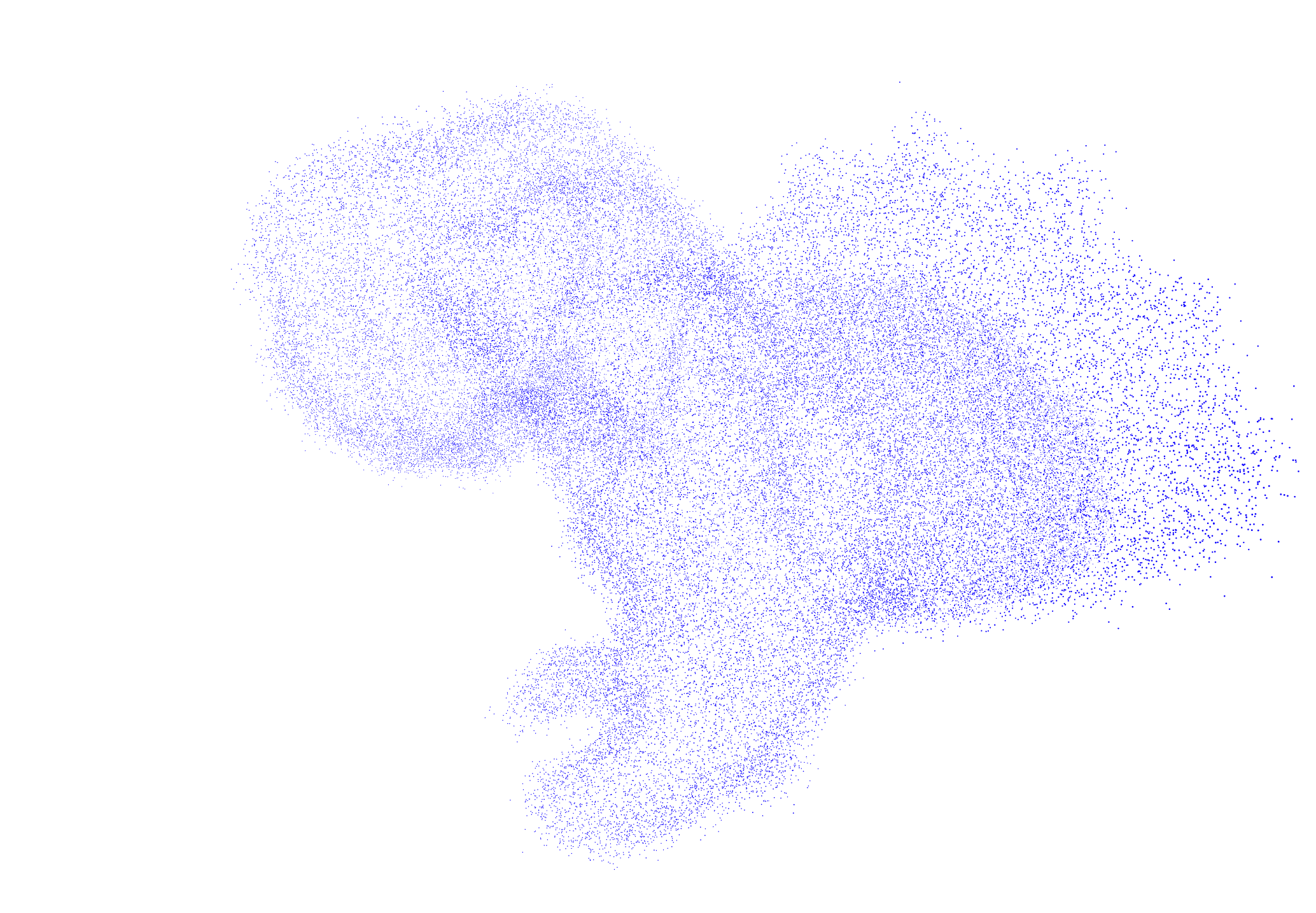}}\!\!\!
	\subfloat[]{\includegraphics[width=0.25\textwidth]{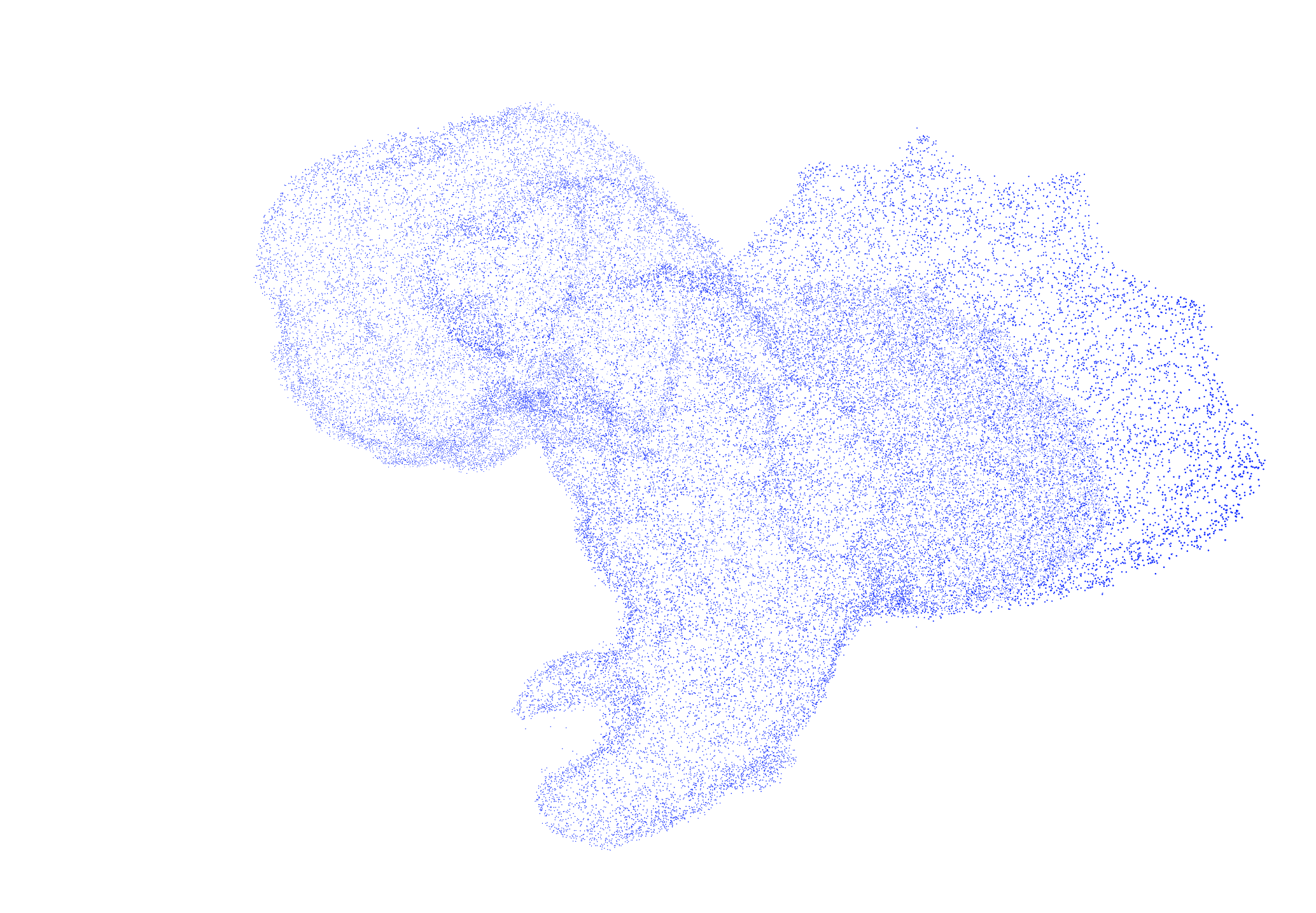}}
	\caption{A visualization of the TUDE algorithm denoising result on the \emph{Gargoyle} model ($\sigma=0.1$). (a) the true model; (b) the noisy model; (c) the denoised model by TUDE algorithm.} 
	\label{figure-gargoyle}
\end{figure}

The experimental results are shown in \Cref{table-results-0.04,table-results-0.05,table-results-0.08,table-results-0.1}, where ``---" means that the MSE does not decrease after the algorithm. A visualization of the TUDE algorithm denoising result on the \emph{Gargoyle} model ($\sigma=0.1$) is shown in \Cref{figure-gargoyle}, where we can see that the denoised model is more compact and regular than the noisy one.


\begin{table}[h!]
	\centering
	\small
	\caption{MSE for different models ($\sigma=0.04$).}
	\label{table-results-0.04}
	\begin{tabular}{l l l l l l}
		\toprule
		\text{Model} &$\text{Noisy}$ & $\text{APSS}$ & $\text{RIMLS}$ &  $\text{GLR}$ & \text{TUDE}\\
		\midrule
		\text{Gargoyle}       & 0.367& 0.258 & 0.275  &   0.251&  0.283\\
		\midrule
		\text{DC}       &  0.338 &  0.227 &  0.245 &   0.217 &  0.248 \\
		\midrule
		\text{Daratech}  & 0.348 &  0.264 &  0.284  &  0.269  &  0.301 \\
		\bottomrule
	\end{tabular}
\end{table}

\begin{table}[h!]
	\centering
	\small
	\caption{MSE for different models ($\sigma=0.05$).}
	\label{table-results-0.05}
	\begin{tabular}{l l l l l l}
		\toprule
		\text{Model} &$\text{Noisy}$ & $\text{APSS}$ & $\text{RIMLS}$ &  $\text{GLR}$ & \text{TUDE}\\
		\midrule
		\text{Gargoyle}       & 0.413& 0.281&  0.298  &  0.269 &  0.301\\
		\midrule
		\text{DC}       &   0.381 & 0.248&   0.266  &  0.231 &  0.265 \\
		\midrule
		\text{Daratech}  &   0.387 &0.328   &  0.363 & 0.308 & 0.331 \\
		\bottomrule
	\end{tabular}
\end{table}

\begin{table}[h!]
	\centering
	\small
	\caption{MSE for different models ($\sigma=0.08$).}
	\label{table-results-0.08}
	\begin{tabular}{l l l l l l}
		\toprule
		\text{Model} &$\text{Noisy}$ & $\text{APSS}$ & $\text{RIMLS}$ &  $\text{GLR}$ & \text{TUDE}\\
		\midrule
		\text{Gargoyle}       &  0.539 & 0.393 & 0.432  &  0.348  &   0.367\\
		\midrule
		\text{DC}       &   0.503 & 0.377 & 0.409  &    0.317   &  0.341\\
		\midrule
		\text{Daratech}  &    0.482  &0.458  & ----   &  0.384  &  0.388\\
		\bottomrule
	\end{tabular}
\end{table}

\begin{table}[h!]
	\centering
	\small
	\caption{MSE for different models ($\sigma=0.1$).}
	\label{table-results-0.1}
	\begin{tabular}{l l l l l l }
		\toprule
		\text{Model} &$\text{Noisy}$ & $\text{APSS}$ & $\text{RIMLS}$ &  $\text{GLR}$ & \text{TUDE}\\
		\midrule
		\text{Gargoyle}       &  0.619& 0.531 &0.583  &  0.436 &  0.416\\
		\midrule
		\text{DC}       &  0.577 & 0.514 &   0.556  &  0.400 &  0.392 \\
		\midrule
		\text{Daratech}  &     0.531& ---- &  ---- &  0.437 &  0.418\\
		\bottomrule
	\end{tabular}
\end{table}

\subsection{Discussions}

We first make some comparisons based on  \Cref{table-results-0.04,table-results-0.05,table-results-0.08,table-results-0.1}.
(1) Compared with the APSS algorithm, TUDE algorithm has better results when $\sigma=0,08, 0.1$, and APSS is better when $\sigma=0.04, 0.05$.
(2) Compared with the RIMLS algorithm, TUDE algorithm has better results when $\sigma=0.08, 0.1$, close results when $\sigma=0.05$ and RIMLS is better when $\sigma=0.04$. 
(3) Compared with the state-of-the-art GLR algorithm, TUDE algorithm has better results when $\sigma=0.1$, and GLR is better when $\sigma=0.04, 0.05, 0.08$. 
Then we can summarize that the proposed TUDE algorithm has better results when tackling the point cloud denoisng problem with high noise.


This is because of the following facts: (1) The APSS and RIMLS algorithms are both based on the local reference domains of the noisy points, which are influenced more heavily by the high noise. 
However, the GLR and TUDE algorithms belong to Non-local category, and thus still work well with high noise.
(2) The GLR algorithm uses projection on the reference plane to find the correspondence \cite{zeng20183d}, which may make it more sensitive to the high noise than the TUDE algorithm.
(3) In the TUDE algorithm, after the denoising process by the compression of Tucker decomposition (i.e., higher order PCA), we continue to manipulate the hard thresholding on the core tensor. In the experiments, we find that the denoising result is much better than just using the Tucker decomposition. The combination of these two processes may be also an important reason why it works better in high noise case.

%

The proposed TUDE algorithm is also competitive in the speed. In fact, 
the running time of each experiment is $4\sim 6$ minutes on the \emph{Daratech} model, and $8\sim 10$ minutes on the \emph{Gargoyle} and \emph{DC} models. 
This is because of that (1) TUDE algorithm is not an iterative algorithm, and (2) the main procedure in TUDE algorithm is the HOOI algorithm, which is of high efficiency \cite{phan2014fast}. 

Now we make some discussions about the parameters of the proposed TUDE algorithm. (1) There is a strong positive correlation between the patch matrix size $\text{K}$ and the Gaussian noise level. This is reasonable as we need larger surface patches when the noise is higher.
(2) We always set N$_{\rm reg}=$ 20 in the experiments. In fact, the final MSE value will be better if we increase the N$_{\rm reg}$ value. However, this would make the algorithm slower. In these three models, the speed would be too slow (more than one hour) if N$_{\rm reg}$ is more than 200.
(3) We always set $\delta_{\rm thre}=0.1$ in the experiments. In fact, on the \emph{Daratech} model, we find that the result will be better, if $\delta_{\rm thre}$ is set to be smaller. One possible reason is that the \emph{Daratech} model has more flat surfaces, while other two models have more round surfaces.

In the experiments, we find that the hard thresholding process sometimes keeps only one entry (with the largest absolute value), and eliminates all other entries. In this case, it is natural to ask whether we could get the same result if we take $(r_1,r_2,r_3) = (1,1,1)$ directly, which is corresponding to the \emph{best rank-1 approximation} \cite{de2000best}. In fact, by the experiments, we find that this is not the case. Taking $(r_1,r_2,r_3) = (1,1,1)$ directly generally gets worse results than the approach of this paper. In other words, combining the Tucker decomposition and hard thresholding is better.

\section{Conclusion}\label{sec-conclusion}

In this paper, we propose a new point cloud denoising algorithm by using the tensor Tucker decomposition to exploit the
self-similarities between local surface patches in a point cloud.
After calculating the Tucker decomposition by the HOOI algorithm, we manipulate the hard thresholding on the compressed tensor to remove the noise.
Finally all the points are placed back, and the average is taken if there are more than one estimators overlapped. From the experiments, we show that this new algorithm is competitive in the speed, and outperforms the state-of-the-art GLR algorithm when the additive Gaussian noise is high ($\sigma=$ 0.1), while the GLR algorithm is better in lower noise cases ($\sigma=0.04, 0.05, 0.08$).
In the future work, based on the TUDE algorithm, we will try to construct a nonlinear model to get better denoising results.

\section{Acknowledgement}

The authors would like to thank Gene Cheung for providing the point cloud models and Matlab codes of GLR algorithm, and Hei Victor Cheng for valuable discussions about the algorithm setting.

\bibliographystyle{IEEE}
\bibliography{HOSVD_PC}

\begin{thebibliography}{10}

\bibitem{li2015hyperspectral}
Chang Li, Yong Ma, Jun Huang, Xiaoguang Mei, and Jiayi Ma,
\newblock ``Hyperspectral image denoising using the robust low-rank tensor
  recovery,''
\newblock {\em J. Opt. Soc. Am. A}, vol. 32, no. 9, pp. 1604--1612, 2015.

\bibitem{lozes2015pde}
Fran{\c{c}}ois Lozes, Abderrahim Elmoataz, and Olivier L{\'e}zoray,
\newblock ``Pde-based graph signal processing for 3d color point clouds:
  Opportunities for cultural heritage,''
\newblock vol. 32, no. 4, pp. 103--111, 2015.

\bibitem{sugimoto2017trends}
Kazuo Sugimoto, Robert~A Cohen, Dong Tian, and Anthony Vetro,
\newblock ``Trends in efficient representation of 3d point clouds,''
\newblock in {\em Proc. APSIPA ASC}, 2017, pp. 364--369.

\bibitem{zhang1998thresholding}
Xiao-Ping Zhang and M.~D. {Desai},
\newblock ``Adaptive denoising based on sure risk,''
\newblock {\em IEEE Signal Processing Letters}, vol. 5, no. 10, pp. 265--267,
  Oct 1998.

\bibitem{zhang1998thresholdingB}
Xiao-Ping Zhang and M.~D. {Desai},
\newblock ``Nonlinear adaptive noise suppression based on wavelet transform,''
\newblock in {\em Proc. ICASSP98}, May 1998, vol.~3, pp. 1589--1592.

\bibitem{zhang1999thresholding}
Xiao-Ping Zhang and Zhi-Quan Luo,
\newblock ``A new time-scale adaptive denoising method based on wavelet
  shrinkage,''
\newblock in {\em Proc. ICASSP99}, March 1999, vol.~3, pp. 1629--1632.

\bibitem{zhang2001thresholding}
Xiao-Ping Zhang,
\newblock ``Thresholding neural network for adaptive noise reduction,''
\newblock {\em IEEE transactions on neural networks}, vol. 12, no. 3, pp.
  567--584, 2001.

\bibitem{zhang2001thresholdingB}
Xiao-Ping Zhang,
\newblock ``Space-scale adaptive noise reduction in images based on
  thresholding neural network,''
\newblock in {\em Proc. ICASSP2001}, May 2001, vol.~3, pp. 1889--1892.

\bibitem{beheshti2010noise}
Soosan Beheshti, Masoud Hashemi, Xiao-Ping Zhang, and N.~Nikvand,
\newblock ``Noise invalidation denoising,''
\newblock {\em IEEE Trans. Sig. Proc.}, vol. 58, no. 12, pp. 6007--6016, 2010.

\bibitem{han2017review}
Xian-Feng Han, Jesse~S Jin, Ming-Jie Wang, Wei Jiang, Lei Gao, and Liping Xiao,
\newblock ``A review of algorithms for filtering the 3d point cloud,''
\newblock {\em Signal Processing: Image Communication}, vol. 57, pp. 103--112,
  2017.

\bibitem{zeng20183d}
Jin Zeng, Gene Cheung, Michael Ng, Jiahao Pang, and Cheng Yang,
\newblock ``3d point cloud denoising using graph laplacian regularization of a
  low dimensional manifold model,''
\newblock {\em arXiv:1803.07252}, 2018.

\bibitem{guennebaud2007algebraic}
Ga{\"e}l Guennebaud and Markus Gross,
\newblock ``Algebraic point set surfaces,''
\newblock {\em ACM Transactions on Graphics (TOG)}, vol. 26, no. 3, pp. 23,
  2007.

\bibitem{oztireli2009feature}
A~Cengiz {\"O}ztireli, Gael Guennebaud, and Markus Gross,
\newblock ``Feature preserving point set surfaces based on non-linear kernel
  regression,''
\newblock {\em Computer Graphics Forum}, vol. 28, no. 2, pp. 493--501, 2009.

\bibitem{osher2017low}
Stanley Osher, Zuoqiang Shi, and Wei Zhu,
\newblock ``Low dimensional manifold model for image processing,''
\newblock {\em SIAM J. on Imaging Sci.}, vol. 10, no. 4, pp. 1669--1690, 2017.

\bibitem{gui2017video}
Lihua Gui, Gaochao Cui, Qibin Zhao, Dongsheng Wang, Andrzej Cichocki, and
  Jianting Cao,
\newblock ``Video denoising using low rank tensor decomposition,''
\newblock in {\em Proc. ICMV 2016}. SPIE, 2017, vol. 10341.

\bibitem{kolda2009tensor}
Tamara~G Kolda and Brett~W Bader,
\newblock ``Tensor decompositions and applications,''
\newblock {\em SIAM Review}, vol. 51, no. 3, pp. 455--500, 2009.

\bibitem{comon2014tensors}
Pierre Comon,
\newblock ``Tensors: a brief introduction,''
\newblock {\em IEEE Signal Proc. Mag.}, vol. 31, no. 3, pp. 44--53, 2014.

\bibitem{rajwade2013image}
Ajit Rajwade, Anand Rangarajan, and Arunava Banerjee,
\newblock ``Image denoising using the higher order singular value
  decomposition,''
\newblock {\em IEEE Trans. on PAMI}, vol. 35, no. 4, pp. 849--862, 2013.

\bibitem{shahid2017scalable}
Nauman Shahid,
\newblock {\em Scalable Low-rank Matrix and Tensor Decomposition on Graphs},
\newblock Ph.D. thesis, Ecole Polytechnique F{\'e}d{\'e}rale de Lausanne, 2017.

\bibitem{gavili2017shift}
Adnan Gavili and Xiao-Ping Zhang,
\newblock ``On the shift operator, graph frequency, and optimal filtering in
  graph signal processing,''
\newblock {\em IEEE Transactions on Signal Processing}, vol. 65, no. 23, pp.
  6303--6318, 2017.

\bibitem{tucker1966some}
Ledyard~R Tucker,
\newblock ``Some mathematical notes on three-mode factor analysis,''
\newblock {\em Psychometrika}, vol. 31, no. 3, pp. 279--311, 1966.

\bibitem{de2000best}
Lieven De~Lathauwer, Bart De~Moor, and Joos Vandewalle,
\newblock ``On the best rank-1 and rank-(r1, r2,..., rn) approximation of
  higher-order tensors,''
\newblock {\em SIAM J on Matrix Ana. \& App.}, vol. 21, no. 4, pp. 1324--1342,
  2000.

\bibitem{rosman2013patch}
Guy Rosman, Anastasia Dubrovina, and Ron Kimmel,
\newblock ``Patch-collaborative spectral point-cloud denoising,''
\newblock in {\em Computer Graphics Forum}. Wiley Online Library, 2013,
  vol.~32, pp. 1--12.

\bibitem{cignoni2008meshlab}
Paolo Cignoni, Marco Callieri, Massimiliano Corsini, Matteo Dellepiane, Fabio
  Ganovelli, and Guido Ranzuglia,
\newblock ``Meshlab: an open-source mesh processing tool,''
\newblock in {\em Eurographics Italian chapter conf.}, 2008, pp. 129--136.

\bibitem{mattei2017point}
Enrico Mattei and Alexey Castrodad,
\newblock ``Point cloud denoising via moving rpca,''
\newblock {\em Computer Graphics Forum}, vol. 36, no. 8, pp. 123--137, 2017.

\bibitem{phan2014fast}
Anh-Huy Phan, Andrzej Cichocki, and Petr Tichavsky,
\newblock ``On fast algorithms for orthogonal tucker decomposition,''
\newblock in {\em IEEE ICASSP 2014}, pp. 6766--6770.

\end{thebibliography}

\end{document}